# A Simple Approach for Finding the Globally Optimal Bayesian Network Structure

**Tomi Silander and Petri Myllymäki**
Complex Systems Computation Research Group (CoSCo)
Helsinki Institute for Information Technology (HIIT)
P.O.Box 68 (Department of Computer Science)
FIN-00014 University of Helsinki, Finland

## Abstract

We study the problem of learning the best Bayesian network structure with respect to a decomposable score such as BDe, BIC or AIC. This problem is known to be NP-hard, which means that solving it becomes quickly infeasible as the number of variables increases. Nevertheless, in this paper we show that it is possible to learn the best Bayesian network structure with over 30 variables, which covers many practically interesting cases. Our algorithm is less complicated and more efficient than the techniques presented earlier. It can be easily parallelized, and offers a possibility for efficient exploration of the best networks consistent with different variable orderings. In the experimental part of the paper we compare the performance of the algorithm to the previous state-of-the-art algorithm. Free source-code and an online-demo can be found at http://b-course.hiit.fi/bene.

## 1 INTRODUCTION

Inspired by Koivisto & Sood (2004), we set ourselves to implement an exact Bayesian network structure discovery for complete discrete data. In its beautiful generality, the original paper concentrates on calculating the probability of any modular feature in Bayesian networks, and hints to the actual structure discovery as a straight forward modification of the general theory. Much of the theory and experiments in Koivisto & Sood (2004) also deals with the cases subject to some given upper bound for the number of parents in the Bayesian network.

In this paper we give a much simpler algorithm for finding the globally optimal Bayesian network structure without any structural constraints. The algorithm itself could deal with structural constraints, but that would complicate the presentation. The simplicity of our method will also reveal obvious ways to distribute the computation to many computers. However, in this paper, we will not concentrate on the distributed implementation.

We hope a simple account of our method together with the freely available software, will rapidly add this technique to the commonly used data analysis toolbox. Furthermore, having the tool handy would hopefully incite further studies on structure discovery algorithms, different scoring metrics, etc. One could also wish for healthy competition on the task.

Finding the best Bayesian network structure is NP-hard (Chickering, Meek, & Heckerman, 2003), so the techniques presented will be feasible only for small networks. We present a best network for a data set with 29 variables, and we conduct our experiments with 30 data sets with at most 20 variables.

After implementing our method, we learned about the work of of Singh and Moore (2005). They had come to similar conclusions about the methods presented in Koivisto & Sood (2004) and attacked the problem of structure learning more directly. However, they state their method, the SM-algorithm, to be feasible for $n < 26$ and demonstrate it with $n = 22$. We claim feasibility for $n < 33$ and demonstrate with $n = 29$. Our method is even simpler than that of Singh and Moore who deploy an intricate mathematical theorem to prevent unnecessary computations. In the experimental part of the paper we demonstrate that our method compares favorably to the SM-algorithm, that we believe to be (former) state-of-the-art. As we will show, by not avoiding any computations we can quickly find the best network consistent with any given variable ordering.

The rest of the paper is organized as follows: In Chapter 2 we will briefly introduce the Bayesian networks and other theoretical concepts on which we will build our approach. The actual method and the algorithms are explained in Chapter 3. After demonstrating the capability of our approach in Chapter 4, we end with some discussion and ideas for future work in Chapter 5.

## 2 DEFINITIONS

### 2.1 BAYESIAN NETWORKS

Bayesian network structures (see, e.g., (Heckerman, 1996)), or simply networks, for a variable set $V = \{V_1, \ldots, V_n\}$ are directed acyclic graphs (DAGs) with exactly one node per variable. We often equate the $V$ by the indices of its variables $V = \{1, \ldots, n\}$. A network $G$ can be described as a vector $G = (G_1, \ldots, G_n)$ of parent sets: $G_i$ is the subset of $V$ from which there are arcs to $V_i$. For example, the network $G = (\{4\}, \{1,3\}, \{1\}, \{\})$ corresponds to the DAG with arcs $\{4 \to 1, 1 \to 2, 3 \to 2, 1 \to 3\}$. We will also use a concept of variable ordering. Ordering of the variable set $V$ is simply the variables of $V$ in some order, e.g. $ord = (4, 1, 3, 2)$. We denote the $i^{th}$ element in this ordering by $ord_i$, i.e., in the example above $ord_2 = 1$. The Bayesian network $G = (G_1, \ldots, G_n)$ is said to be consistent with an ordering $ord$ when for all $i$, $G_i \subseteq \bigcup_{j=1}^{i-1}\{ord_j\}$, i.e., when all the parents of the node precede the node in the ordering. Our example network $G$ above is consistent with our example ordering $ord$.

There is one crucial fact about DAGs we will frequently in the sequel: every DAG has at least one node with no outgoing arcs, so at least one node is not a parent of any other node. These nodes are called *sinks* of the network. The existence of a sink follows from the fact that, due to the acyclicity, every directed path in a DAG must have a finite length, and the end of a maximally long directed path is necessarily a sink.

### 2.2 DATA

Bayesian networks are usually used as statistical models for data. In this paper we concentrate on multivariate discrete data where we have many variables $V = \{V_1, \ldots, V_n\}$, and each variable $V_i$ has a (usually very) finite number of values $(v_{i_1}, \ldots, v_{i_{n_i}})$. The variables are measured in nominal scale, i.e. the ordering of the values is irrelevant. We often refer to the values by their indices, so that the values of $V_i$ are taken to be $(1, \ldots, n_i)$.

The complete discrete data $D = (D_1, \ldots, D_N)$ is a collection of $N$ data-vectors $D_j = (D_{j1}, \ldots, D_{jn})$ in which the component $D_{ji}$ has one of the values of $V_i$. It is customary to think a data-vector as a row in an $N \times n$ data-matrix in which the columns correspond to the variables.

We also define the projection of a data to a non-empty variable set. The projection of a data-vector $D_j$ to the variable set $W$ means selecting the components belonging to $W$ from the $D_j$. We denote this projection by $D_j^W$. Performing this projection to all the data vectors gives us projected data $D^W$, which corresponds to picking the columns $W$ from the data-matrix.

To summarize the data, we define the contingency table $CT(W)$ to be a list the frequencies of different data-vectors $d^W$ in $D^W$. Our main task, however, will be to calculate conditional frequency tables $CFT(v, W)$ that record how many times different values of the variable $v$ occur together with different vectors $d^{W-\{v\}}$ in the data.

### 2.3 SCORES

Our goal is to find the best Bayesian network structure for the given complete (no missing values), discrete data. The definition of the "best" is usually formalized by defining a scoring function that, given a data, attaches a real number to any given network. We take the problem to be a maximization problem: the better the network, the greater the score.

The nature of many common scoring functions is such that several different networks may have equal scores (Chickering, 1995). Therefore, we often write "a best" instead of "the best". During the process of finding a best network structure, whenever "a best" anything is selected, it does not matter which one is selected, the end result will still be a best network. It is worth mentioning that, for many popular scores, finding a best network allows us to easily find other best networks.

In order to use the method to be presented, the scoring functions have to be modular, i.e., given the data, the score of a Bayesian network structure $G = (G_1, \ldots, G_n)$ for variables $V = (1, \ldots, n)$ must be decomposable to local scores:

$$score(G) = \sum_{i=1}^{n} score_i(G_i) = \sum_{i=1}^{n} score(CFT(i, G_i)),$$

so that the score of the network is the sum of the local scores that only depend on the conditional frequency table for one variable and its parents. Many popular scores like BDe, BIC and AIC decompose like this (Chickering, 1995). We denote the highest scoring network for a variable set $W$ by $G^*(W)$. We also define $sink^*(W)$ to be the lowest numbered sink of the $G^*(W)$ (actually, any random sink would do for our purposes). Finally, we denote by $ord^*(W)$ an ordering (again, there may be several) of $W$ that is consistent with the best network $G^*(W)$.

The local scoring function $score_i$ measures the goodness of $V_i$'s parents. This idea naturally leads to finding the best parents for a variable $V_i$ in any given parent candidate set $C$:

$$g_i^*(C) = \arg\max_{g \subseteq C} score_i(g).$$

# 3 SIMPLE METHOD FOR FINDING BEST NETWORK STRUCTURES

We are now ready to present our approach for finding the best Bayesian network structure. Our method has five logical steps:

1. Calculate the local scores for all $n2^{n-1}$ different (variable, variable set)-pairs.

2. Using the local scores, find best parents for all $n2^{n-1}$ (variable, parent candidate set)-pairs.

3. Find the best sink for all $2^n$ variable sets.

4. Using the results from Step 3, find a best ordering of the variables.

5. Find a best network using results computed in Steps 2 and 4.

Step 3 of the algorithm is based on the following observation: The best network $G^*(W)$ must have a sink $s$, and that sink must have incoming arcs from its best possible parents $g_s^*(W \setminus \{s\})$. The rest of the nodes and the arcs must form the best possible network for the variables $W \setminus \{s\}$.

More formally,

$$sink^*(W) = \arg\max_{s \in W} skore(W, s), \quad (1)$$

where

$$skore(W, s) = score_s(g_s^*(W\setminus\{s\})) + score(G^*(W\setminus\{s\})).$$

To see this, let us pick an arbitrary sink node $s$ from the best network $G^*(W)$. Now the subnetwork $G^{-s}$ of $G^*(W)$ from which the node $s$ and the arcs coming from $g_s^*(W \setminus \{s\})$ to it are removed must be a best network for $W\setminus\{s\}$, since if there were a strictly better network $G^\dagger$ for $W \setminus \{s\}$, augmenting $G^\dagger$ with $s$ and $g_s^*$ would yield a network strictly better than $G^*(W)$, which is a contradiction.

Best sinks immediately yield the best ordering in reverse order:

$$ord_i^*(V) = sink^*(V \setminus \bigcup_{j=i+1}^{|V|} \{ord_j^*(V)\}). \quad (2)$$

Having $ord_i^*(V)$ and $g_i^*(W)$ available for all $W \subseteq V$, the step 5 is as easy, since

$$G_{ord_i^*(V)}^* = g_{ord_i^*(V)}^*(V \setminus \bigcup_{j=1}^{i-1}\{ord_j^*(V)\}), \quad (3)$$

so that for the $i^{th}$ variable in the optimal ordering, we simply pick the best parents from its predecessors.

In the following subsections we explain how the steps 1–5 can be accomplished in reasonable time.

## 3.1 CALCULATING LOCAL SCORES

This is the only step in which the data is needed. It is also the most time consuming phase of the algorithm. We start by calculating the contingency table for all the variables $V$ and incrementally (decrementally) calculate contingency tables for smaller variable subsets (Figure 1).

| Data vectors | # | | Data vectors | # |
|---|---|---|---|---|
| 0 1 1 **0** 1 | 3 | | 0 1 1 1 | 6 |
| 0 1 1 **1** 0 | 15 | | 0 1 1 0 | 15 |
| 0 1 1 **1** 1 | 3 | $\Rightarrow$ | 1 0 1 0 | 5 |
| 1 0 1 **0** 0 | 1 | | 1 0 1 1 | 7 |
| 1 0 1 **1** 0 | 4 | | | |
| 1 0 1 **1** 1 | 7 | | | |

Figure 1: Marginalizing a variable (in bold) out of the contingency table yields a smaller contingency table.

From each contingency table, we calculate conditional frequency tables for the variables appearing in the contingency table (Figure 2). These conditional frequency tables can then be used to calculate the local scores.

| Data vectors | # | | Data vectors | (F0,F1) |
|---|---|---|---|---|
| 0 1 1 **1** | 6 | | 0 1 1 | (15, 6) |
| 0 1 1 **0** | 15 | $\Rightarrow$ | 1 0 1 | ( 5, 7) |
| 1 0 1 **0** | 5 | | | |
| 1 0 1 **1** | 7 | | | |

Figure 2: Building a conditional frequency table for a variable (on bold) in a contingency table.

The main procedure, $GetLocalScores$, (Algorithm 1) is called with a contingency table $ct$ and the variables $evars$ to be marginalized from it. Initially, it is called with a contingency table for all the variables and the whole variable set $V$ as $evars$. The algorithm is simply a depth first traversal of smaller and smaller contingency tables.

The pseudocode of the Algorithm 1 assumes some helper functions: $Vars(ct)$ returns the set of variables in the contingency table $ct$, $Ct2ct(ct, v)$ produces a contingency table by marginalizing the variable $v$ out of $ct$ as demonstrated in Figure 1, $Ct2cft(ct, v)$ yields a conditional frequency table as shown in Figure 2, and finally, the function $Score(cft)$ calculates the local score based on the conditional frequency table. The algorithm produces a mapping $LS$ from (variable, parent set)-pairs to the corresponding local scores.

---

**Algorithm 1** $GetLocalScores(ct, evars)$

/* Turn $ct$ to $cfts$ and further to local scores */
**for all** $v \in Vars(ct)$ **do**
　　$LS[v][Vars(ct) \setminus \{v\}] \leftarrow Score(Ct2cft(ct, v))$
**end for**

/* Recursively call $GetLocalScores$ */
**if** $|Vars(ct)| > 1$ **then**
　　**for all** $v \in evars$ **do**
　　　　$GetLocalScores(Ct2ct(ct, v), \{1, \cdots, v-1\})$
　　**end for**
**end if**

---

As an implementational note, the local scores do not have to be kept in the memory, but can be stored on the disk as soon as they are produced. Therefore, the memory requirement of this first phase is about $|V|$ times the size of the initial contingency table.

This procedure also serves as a basis for parallelizing the algorithm. By initially calling the $GetLocalScores$ with different contingency tables, the task can be divided to separate subtasks.

### 3.2 FINDING BEST PARENTS

Having calculated the local scores, finding the best parents for a variable $v$ from a set $C$ can be done recursively. The best parents in $C$ for $v$ are either the whole candidate set $C$ itself or the best parents for $v$ from one of the smaller candidate sets $\{C \setminus \{c\} \mid c \in C\}$. More formally,

$$score_i(g_i^*(C)) = \max(score_i(C), score1(C)), \quad (4)$$

where

$$score1(C) = \max_{c \in C} score_i(g_i^*(C \setminus \{c\}))).$$

This translates directly into an algorithm that goes through all the candidate sets in lexicographic order, and evaluates the formula above for each of them. Below, you can find the pseudocode of the Algorithm 2 that takes the variable set $V$, the variable $v$ and the previously calculated local scores $LS$, and finds the best parent set $bps$ for all the possible parent candidate sets of $v$.

---

**Algorithm 2** $GetBestParents(V, v, LS)$

$bps$ = array 1 to $2^{|V|-1}$ of variable sets
$bss$ = array 1 to $2^{|V|-1}$ of local scores
**for all** $cs \subseteq V \setminus \{v\}$ in lexicographic order **do**
　　$bps[cs] \leftarrow cs$
　　$bss[cs] \leftarrow LS[v][cs]$
　　**for all** $cs1 \subset cs$ such that $|cs \setminus cs1| = 1$ **do**
　　　　**if** $bss[cs1] > bss[cs]$ **then**
　　　　　　$bss[cs] \leftarrow bss[cs1]$
　　　　　　$bps[cs] \leftarrow bps[cs1]$
　　　　**end if**
　　**end for**
**end for**
**return** $bps$

---

A direct calculation shows that the algorithm runs in time $o((n-1)2^{n-2})$. It has to be run for each variable which naturally can be done in parallel. The algorithm requires $2^{n-1}$ scores and parent sets to be held in memory.

### 3.3 FINDING BEST SINKS

As mentioned before, a best network for a variable set $W$ can always be constructed by first finding a best sink $s$ in $W$, then constructing the best network for $W \setminus \{s\}$, and finally picking the best parents for $s$ from the variables in $W \setminus \{s\}$.

---

**Algorithm 3** $GetBestSinks(V, bps, LS)$

**for all** $W \subseteq V$ in lexicographic order **do**
　　$scores[W] \leftarrow 0.0$
　　$sinks[W] \leftarrow -1$
　　**for all** $sink \in W$ **do**
　　　　$upvars \leftarrow W \setminus \{sink\}$
　　　　$skore \leftarrow scores[upvars]$
　　　　$skore \leftarrow skore + LS[sink][bps[sink][upvars]]$
　　　　**if** $sinks[W] = -1$ **or** $skore > scores[W]$ **then**
　　　　　　$scores[W] \leftarrow skore$
　　　　　　$sinks[W] \leftarrow sink$
　　　　**end if**
　　**end for**
**end for**
**return** $sinks$

---

Again, this idea translates directly into an algorithm (Algorithm 3). We go through all the variable sets, and from each set $W$ we select a sink that yields a best network for $W$. To make this selection, the only thing we need is the scores of the best networks for $W$'s subsets and the local score of each sink candidate given its best parents in $W$. The latter information is calculated in Step 2. By going through the variable sets in lexicographic order we can guarantee that, for every set, the scores of the best networks for its subsets are already calculated.

The algorithm runs in time $o(n2^{n-1})$ and keeps $2^n$ scores in memory. The sinks are not needed during the algorithm, thus they can be stored one by one as soon as they are produced.

### 3.4 FINDING BEST ORDERING

The procedure to find the best ordering is simply a non-recursive implementation of the equation (2).

---
**Algorithm 4** $Sinks2ord(V, sinks)$

$ord$ = array 1 to $|V|$ of variables
$left = V$
**for** $i = |V|$ **to** 1 **do**
  $ord[i] \leftarrow sinks[left]$
  $left \leftarrow left \setminus \{ord[i]\}$
**end for**
**return** $ord$

---

### 3.5 FINDING BEST NETWORK

After calculating the best parents, $bps$, for all the (variable, parent candidate set)-pairs, as explained in subsection 3.2, we can use equation (3) to quickly find a best network consistent with any given ordering of the variables:

---
**Algorithm 5** $Ord2net(V, ord, bps)$

$parents$ = array 1 to $|V|$ of variable sets
$predecs \leftarrow \emptyset$
**for** $i = 1$ **to** $|V|$ **do**
  $parents[i] \leftarrow bps[ord[i]][predecs]$
  $predecs \leftarrow predecs \cup \{ord[i]\}$
**end for**
**return** $parents$

---

## 4 EXPERIMENTS

To demonstrate the capability of our method, we picked 30 publicly available data sets and conducted series of experiments that would have been difficult to perform without our software. As the main purpose of these experiments was just to illustrate what type of problems can be studied with this software, more elaborate empirical studies were left as future work.

### 4.1 STUDYING MAXIMAL IN-DEGREES

The search for the best network would be feasible for a much larger number of variables, if we could safely set an upper limit to the maximum number of parents (in-degree) any node can have. We used our 30 data sets to study the issue. For each of the data sets, we first constructed a best net using BDe score (with equivalent sample size 1.0) and a best net using the BIC score. The maximal in-degrees ($D_1$ for BDe and $D_2$ for BIC) are reported in Table 1. We also list the number of variables $n$, number of data vectors $N$, and the running times, $T_1$ and $T_2$, required to find the best networks using BDe and BIC scores respectively. All the experiments were run on 2.20Mhz Compaq Evo N800w laptop having 1GB of memory and running Linux 2.6.

Table 1: Maximum in-degrees with BDe(1) and BIC(2) scores.

| Data | n | N | $T_1$ | $T_2$ | $D_1$ | $D_2$ |
|---|---|---|---|---|---|---|
| balance | 5 | 625 | 0 | 0 | 1 | 1 |
| iris | 5 | 150 | 0 | 0 | 2 | 1 |
| thyroid | 6 | 215 | 0 | 0 | 2 | 1 |
| liver | 7 | 345 | 0 | 0 | 1 | 1 |
| ecoli | 8 | 336 | 0 | 0 | 2 | 1 |
| abalone | 9 | 4177 | 0 | 0 | 3 | 2 |
| diabetes | 9 | 768 | 0 | 0 | 1 | 1 |
| post-op | 9 | 90 | 0 | 0 | 0 | 0 |
| yeast | 9 | 1484 | 0 | 0 | 1 | 1 |
| bc | 10 | 286 | 0 | 0 | 1 | 1 |
| shuttle | 10 | 58000 | 0 | 0 | 5 | 3 |
| tic-tac | 10 | 958 | 1 | 0 | 3 | 3 |
| bc-wisc | 11 | 699 | 0 | 0 | 2 | 2 |
| glass | 11 | 214 | 0 | 0 | 4 | 1 |
| pg-block | 11 | 5473 | 0 | 0 | 3 | 2 |
| heart-cl | 14 | 303 | 5 | 4 | 1 | 1 |
| heart-hu | 14 | 294 | 4 | 3 | 4 | 4 |
| heart-st | 14 | 270 | 5 | 4 | 1 | 1 |
| wine | 14 | 178 | 3 | 3 | 2 | 2 |
| adult | 15 | 32561 | 334 | 314 | 4 | 4 |
| aus | 15 | 690 | 10 | 7 | 2 | 2 |
| credit | 16 | 690 | 32 | 24 | 2 | 2 |
| letter | 17 | 20000 | 603 | 523 | 5 | 4 |
| voting | 17 | 435 | 77 | 45 | 3 | 3 |
| zoo | 17 | 101 | 18 | 11 | 7 | 2 |
| tumor | 18 | 339 | 95 | 73 | 3 | 2 |
| lympho | 19 | 148 | 171 | 147 | 13 | 2 |
| vehicle | 19 | 846 | 391 | 288 | 4 | 2 |
| hepatitis | 20 | 155 | 332 | 263 | 13 | 2 |
| segment | 20 | 2310 | 774 | 387 | 9 | 17 |

We notice that while the maximum in-degree is usually low, this is not always the case. For example, in Koivisto & Sood (2004) the best network for the zoo-data was learned using the maximum in-degree bound of 6, while the best network actually has the maximum in-degree of 7. Furthermore, it took 42 seconds for them to find the best network with the in-degree bound, while in our case it took only 18 seconds to find a best network without any in-degree bound. While setting the in-degree bound may prevent finding the score maximizing network, it may still be that the best network structure within the bounds is good enough (or even better) for specific tasks like prediction. This is one of the future studies that is made possible with our new tool.

## 4.2 SPEED COMPARISON

We also compared the running time ($T_{us}$) of our algorithm to that of the state-of-the-art SM-algorithm ($T_{SM}$). We were able to use four same datasets as Singh & Moore (2005). The results are presented in Table 2. While some of the differences may be due to the different computers used and possibly some other implementational details, the results clearly show that our method is competitive. The comparison also confirmed that our algorithm and its implementation is probably correct, since the BDe scores obtained for these data sets were exactly the same.

Table 2: Comparison of our method and SM.

| Data | n | N | $T_{SM}$ | $T_{us}$ |
|---|---|---|---|---|
| nursery | 9 | 12960 | 10 | 1 |
| parity | 10 | 1000000 | 650 | 4 |
| adult | 15 | 10000 | 4050 | 168 |
| letters | 17 | 20000 | 13146 | 532 |

The faster time for adult-data in this table compared to the Table 1 is due to the fact that in previous runs we used double precision floating point numbers for scores, just in case BIC needs them, while in this run we used single precision floating point numbers.

## 4.3 STUDYING THE ROLE OF ESS

As a pretaste of a more meaningful study, we learned the best network for the yeast data (9 variables, 1484 data vectors) with different values of the equivalent sample size (ESS). The results in Figure 3 show that by going form very small values of ESS (2e-20) to very large values of ESS (34000), it is possible to get any number of arcs to your "globally optimal" network structure. For this data set, the network learned with BIC has 6 arcs, which corresponds to the ESS value of about 0.02. The results clearly suggests further studies about the robustness of the MAP structure.

## 4.4 STUDYING PREDICTION

As an example of the possibility to study the predictive behaviour of the score maximizing network structures,

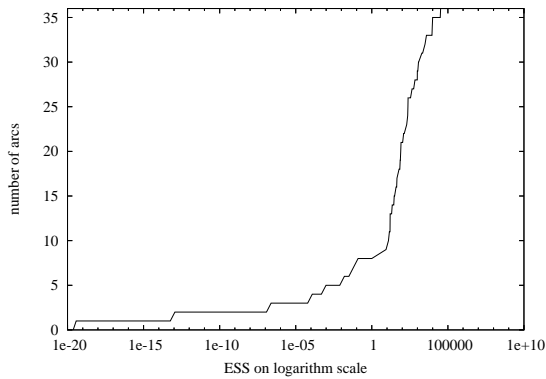

Figure 3: Number of arcs in the optimal network as a function of ESS.

we started by learning a best network for the shuttle data (10 variables, 58000 data vectors) using BDe score with ESS of 1.0. We then used this network with its expected parameter values to generate 100 training samples of sample sizes between 10 and 50000, and tested the predictive performance of the best network structures learned with BDe-score (ESS=1.0) and BIC-score. The predictive performance was taken to be the average marginal probability of the hundred freshly generated test vectors as predicted by the network structures with parameters integrated out (using the ESS of 1.0 again). We also included the network structure of the data generating model to serve as a reference.

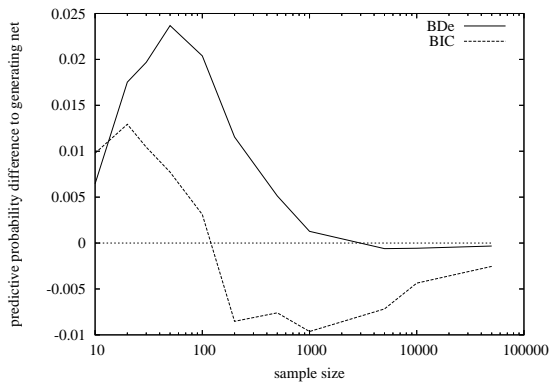

Figure 4: Average predicted probability difference to the network of the generating model as a function of the sample size.

Not surprisingly, the results (Figure 4) show that the network structure of the generating model is too com-plex for the small sample sizes, for which the conservative BIC score yields the best results. However, already with sample size 20 BDe score beats BIC score. Eventually, the right network structure overcomes the sampling variance and passes the BIC and BDe optimal networks.

## 4.5 BUILDING A BIG NET

To demonstrate the feasibility of our approach beyond the n=25 limit, the maximum feasibility stated for the SM-algorithm, we built a network of 29 variables (Figure 5). The data used in this experiment contains information about 194 nations and their flags. This data was selected because of its free availability and its suitable number of variables. The flag-data can be downloaded at http://www.ailab.si/orange/datasets.asp.

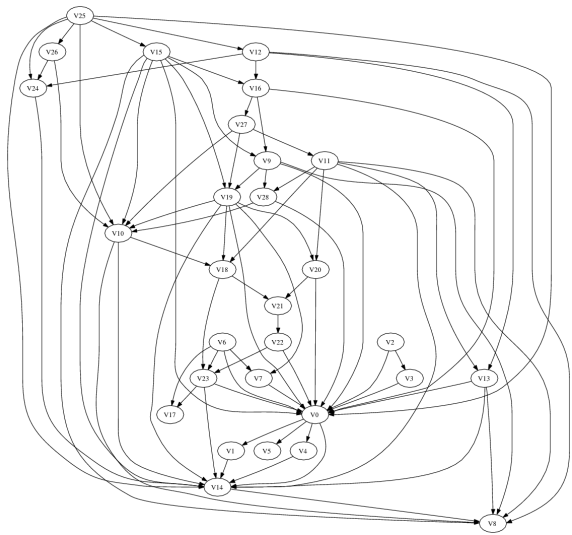

Figure 5: A best network structure for the flag-data.

We started by dividing the calculation of the local scores to 1025 smaller jobs and distributing these jobs to 4 dual-core computers (dual-3GHz Intel Xeon, 512KB L2 cache, 4GB RAM). Calculating the local scores took 6 hours and 16 minutes. Step 2, finding the best parents, took 3 hours, but this task could be parallelized. Step 3, finding the best sinks, took 20 minutes, and Steps 4 and 5 took less than a second each, so the whole task of finding a best 29 variable network was over in about 10 hours. Had we used just one single processor computer, it would have taken us 50 hours to complete the task. We also run stochastic greedy search for two hours, but in that time the search was unable to find the optimal network. The BDe score obtained by the greedy search was 33.7 worse than that of the optimal network. In the future, we plan to conduct a proper study of heuristic search methods.

## 4.6 STUDY OF ORDERINGS

The question about the plausibility of different orderings often arises when we want to speculate about the causal relations between the variables. Therefore, in the last experiment, we used the ability to quickly find the best network for any ordering of the variables.

We picked an optimal ordering of 29 variables constructed in the previous experiment, and evaluated the scores of the best networks consistent with the orderings close to the optimal ordering. We first considered all 28 possible rotations, 14 left and 14 right, of the optimal ordering. The resulting scores of the best networks can be seen in Figure 6. We can clearly see the tendency that the more we rotate the ordering, the worse the networks get. The very first rotation right drops the score by 68 corresponding to the probability ratio of $10^{30}$.

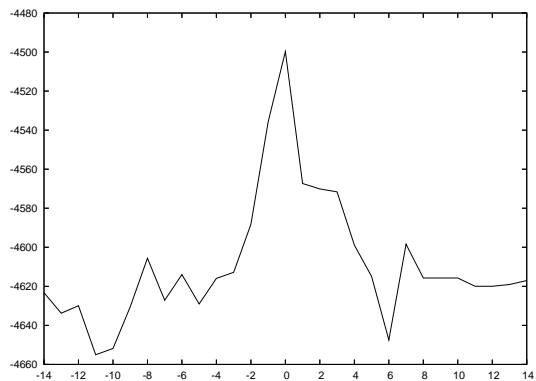

Figure 6: Network scores for rotations.

We also studied $29 \times 28/2 = 392$ possible swappings of two variables in the optimal ordering. The results are presented in Figure 7. Dark colors indicate bad swappings.

It is easy to see how swappings of variables appearing near each other in the optimal ordering (those close to the diagonal) often yield the same optimal score, while swapping variables far from each other often makes the the score worse. One can also notice, for example, that swapping the last variable ($V_8$) with almost any other variable significantly decreases the score. Similarly, the central variable $V_0$ clearly wants to keep its position in a DAG.

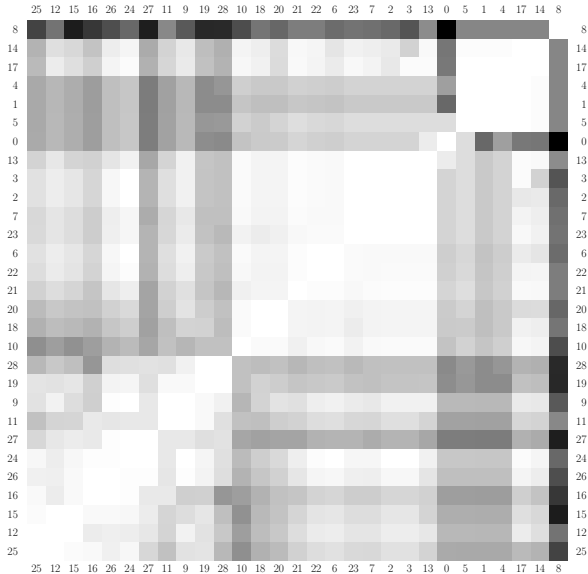

Figure 7: Network scores for swappings.

## 5 DISCUSSION AND FUTURE WORK

We have presented a straightforward method to find best Bayesian networks and demonstrated its feasibility. We believe our method to be the current state-of-the-art.

By the nature of the problem, but also of the method, this approach will not scale up indefinitely. In the current implementation, the memory requirement ($2^{n+2}$ bytes) is the first factor to constrain the size of the networks that can be constructed. Currently, the n=32 is the practical upper limit since it requires over 16GB of memory. Also the disk-space requirement ($12n2^{n-1}$ bytes) can be restrictive. Distributing the computation makes the process faster, but in that case one should ensure that the disk access in a distributed environment is efficient.

However, this is hardly the end of the road. The memory intensive steps 2 and 3 have a very regular memory access pattern, and it is actually possible to get rid of the the memory requirement and turn it to a (bigger) disk-space requirement. We have already implemented this kind of version of the algorithm and we are currently testing it.

In this paper, we have learned the best networks without any constraints to the maximum number of parents. It is easy to see that the algorithm could be modified to benefit from in-degree constraints. In the future, we will study the benefits of in-degree constraints in our method and whether some other approaches can better exploit these constraints.

The computational complexity of our algorithm is $o(n^2 2^{n-2})$, while the complexity of the SM-algorithm is only $o(n2^{n-1})$. It could be possible to drop the computational complexity of our method to $o(n2^{n-1})$ as well, but with small $n$, as it will be due to the nature of the problem, this theoretical result is of little interest. As demonstrated in this paper, the conceptual simplicity of the algorithm that allows an efficient implementation more than compensates the slight (by the factor of n/2) increase in theoretical complexity.

The experiments in the paper are meant to be examples of the kind of studies that can be pursued using this tool. There are many experiments we can think of, but there are certainly many more experiments other people can come up with. Our hope is that by offering the method and its implementation, we can help the researchers to conduct a wider selection of experiments of their interest.

### Acknowledgements

This work was supported in part by the Academy of Finland under the projects Prose and Civi, and by the Finnish Funding Agency for Technology and Innovation under the projects PMMA and SIB, and by the IST Programme of the European Community under the PASCAL Network of Excellence, IST-2002-506778. This publication only reflects the authors' views.